\documentclass{style}

\title{Estimating Human Poses Across Datasets: A Unified Skeleton and Multi-Teacher Distillation Approach}

\addauthor{Muhammad Saif Ullah Khan}{}{1,2}
\addauthor{Dhavalkumar Limbachiya}{}{2}
\addauthor{Didier Stricker}{}{1,2}
\addauthor{Muhammad Zeshan Afzal}{}{1,2}

\addinstitution{
 German Research Center for Artificial Intelligence (DFKI)\\
 Kaiserslautern, Germany
}
\addinstitution{
 University of Kaiserslautern-Landau (RPTU)\\
 Kaiserslautern, Germany
}

\runninghead{Khan et al.}{Estimating Human Poses Across Datasets}

\usepackage{amsfonts}
\usepackage{graphicx}
\usepackage{booktabs}
\usepackage{multirow}

\begin{document}

\maketitle

\begin{abstract}
Human pose estimation is a key task in computer vision with various applications such as activity recognition and interactive systems. However, the lack of consistency in the annotated skeletons across different datasets poses challenges in developing universally applicable models. To address this challenge, we propose a novel approach integrating multi-teacher knowledge distillation with a unified skeleton representation. Our networks are jointly trained on the COCO and MPII datasets, containing 17 and 16 keypoints, respectively. We demonstrate enhanced adaptability by predicting an extended set of 21 keypoints, 4 (COCO) and 5 (MPII) more than original annotations, improving cross-dataset generalization. Our joint models achieved an average accuracy of 70.89 and 76.40, compared to 53.79 and 55.78 when trained on a single dataset and evaluated on both. Moreover, we also evaluate all 21 predicted points by our two models by reporting an AP of 66.84 and 72.75 on the Halpe dataset. This highlights the potential of our technique to address one of the most pressing challenges in pose estimation research and application - the inconsistency in skeletal annotations.
\end{abstract}

\section{Introduction}
\label{sec:intro}

2D human pose estimation using deep learning~\cite{dang2019deep, zhao2021orderly, thai2023multiple, jiang2023rtmpose, zheng2023deep} is crucial for various computer vision applications, such as activity recognition~\cite{song2021human}, augmented reality~\cite{marchand2015pose}, and interactive systems~\cite{liu2022arhpe}. However, many datasets are available~\cite{andriluka14cvpr,lin2014microsoft,li2020pastanet}, each typically featuring a unique set of annotated body joints and skeleton topology~\cite{dang2019deep}. This diversity necessitates the training of separate models for each dataset, which not only restricts the applicability of these models to their respective skeletons but also limits their performance on new, unseen datasets with different skeletons.

To address this critical issue, we introduce a novel framework that unifies human pose estimation across datasets. Our approach employs multi-teacher knowledge distillation~\cite{hinton2015distilling, thai2023multiple} combined with pose union learning to train a single, adaptable student network. This network learns a comprehensive skeleton topology that represents the union of multiple datasets. This significantly improves its generalization and adaptability across various pose estimation tasks. Furthermore, by leveraging a set of pre-trained teacher networks, our method efficiently distills the knowledge from multiple datasets and teacher networks into a single student model. This allows the student model to predict a broader set of keypoints than any individual teacher.

We demonstrate the effectiveness of our proposed approach using multiple variants of the RTMPose model~\cite{jiang2023rtmpose}. We train these models on a unified dataset combining the MPII~\cite{andriluka14cvpr} and COCO~\cite{lin2014microsoft} datasets, containing 16 and 17 annotated body keypoints, respectively. These datasets share 12 common keypoints, with 5 unique points in MPII and 4 unique points in COCO. We train our model to predict a superset of the two skeletons, comprising 21 keypoints in total. We then show the performance of our unified skeleton predictors on the validation sets of both datasets using their respective keypoints. Moreover, we also evaluate the precision of all 21 predicted keypoints using the groundtruth annotations from the Halpe dataset~\cite{li2020pastanet}. This dataset provides 26 body points for images from the COCO validation set. These include all keypoints predicted by our model. Our results indicate that our approach enables the model to learn a unified skeleton that can be generalized to images in both datasets by accurately predicting joints in their images that were not present in their groundtruth annotations. This represents a step forward in addressing the limitations imposed by dataset-specific skeletons and can enable applications that require a more comprehensive understanding of body joints.

To summarize, the contributions of this paper are as follows:

\begin{enumerate}
    \item Introduction of a novel framework that unifies pose estimation across diverse datasets by integrating multi-teacher knowledge distillation and pose union learning, demonstrating competitive performance.
    \item Utilization of the unified model to predict previously unknown keypoints in the COCO and MPII dataset images with high accuracy.
    \item Extensive experimental results to support our approach and validate its effectiveness for pose estimation across the COCO and MPII datasets.
\end{enumerate}

Moreover, our unified model can potentially be used to enhance the ground-truth annotations for these datasets themselves using active learning strategies. While we leave this as future work, our research opens multiple avenues in cross-dataset pose estimation and keypoint dataset enhancement.

The remainder of this paper is organized as follows: Section~\ref{sec:related_work} reviews related work in human pose estimation, knowledge distillation, and cross-dataset pose estimation, providing context for our contributions. Section~\ref{sec:methodology} details our methodology, explaining the architecture and training process of our model. Sections~\ref{sec:experiments} and~\ref{sec:ablations} present our experimental setup, results, and ablation studies, showcasing the empirical evidence supporting our claims. Finally, Section~\ref{sec:conclusions} concludes the paper with a discussion of our findings and their implications for future research in human pose estimation.


\section{Related Work}
\label{sec:related_work}

The development of robust and accurate human pose estimation models and the use of knowledge distillation to enhance training efficiency represent two pivotal research areas in deep learning. This section reviews the current state of the art in these domains and highlights the gaps our work aims to bridge.

\subsection{Human Pose Estimation}

2D human pose estimation using deep learning is a fundamental task in computer vision to locate human joints (keypoints) in images or video frames. The two-stage techniques predominantly fall into either top-down or bottom-up categories. Top-down methods~\cite{li2021tokenpose,xu2022vitpose,khirodkar2021multi,jiang2023rtmpose} first detect individual people in the image and then predict keypoints for each detected person, often leading to higher accuracy but at the cost of computational efficiency. In contrast, bottom-up methods~\cite{cheng2020higherhrnet,osokin2018real,wang2022lite,papandreou2018personlab} detect all keypoints in the image first and then cluster them into individual poses, which can be more scalable but might struggle with accuracy in dense human environments. Alternatively, some recent works~\cite{nie2019single,shi2021inspose,jin2022single,shi2022end,lu2023rtmo} use a one-stage approach using a single model for the complete pose estimation pipeline. Keypoint predictions are typically represented directly as coordinates or, more commonly, through heatmaps, where the latter involves predicting a confidence map for each keypoint and has shown considerable success due to its robustness to pixel-level noise and variability in human appearances. These methods leverage a variety of network architectures, with more recent works using transformers~\cite{xu2022vitpose,li2021tokenpose,sun2024rethinking,shi2022end,zheng20213d}, which have been adapted better to capture the spatial hierarchies inherent in human poses. Despite significant advancements, the challenge remains to enhance the precision and efficiency of these models, especially in unconstrained real-world scenarios where occlusions and complex human interactions are common.

\subsection{Knowledge Distillation}

Knowledge distillation~\cite{hinton2015distilling} is a technique originally developed to compress the knowledge of a large, complex model (teacher) into a smaller, more efficient model (student) with minimal performance loss. Logit-based knowledge distillation~\cite{hinton2015distilling,online_KD6,MultiLevelLD3,online_KD_MB6,ImprovedKD_TA7,blindly_T_KD5,decoupled_KD4,decoupled_KD4} utilizes the soft output probabilities from the teacher model to guide the training of the student model. By employing a loss function that combines temperature-scaled cross-entropy with Kullback-Leibler divergence~\cite{kullback1951information}, the student model is encouraged not only to mimic the teacher’s output distribution but also to achieve correct label predictions on its own. This method effectively transfers the intricate decision boundaries learned by the teacher to the student, enhancing its performance beyond what could be achieved through direct training alone. Feature-based distillation~\cite{info_dist5,refused_TC_KD8,Review5,class_attention_transfer6,overhaul_FD5,kd_pose7,KD_transformer4,KD_seg7,FitNet6,crd8,KD_FMT7} and relation-based distillaion~\cite{KD_STeacher6,KD_detection7,KD_relational3,KD_correlation6,KD_similarity34,SRRL87} have also been proposed to improve the knowledge transfer between teacher and student. Recent advancements, such as the introduction of a teacher assistant model~\cite{mirzadeh2020improved}, have further refined this process by staging the knowledge transfer through an intermediate-sized model, which bridges the gap between the high-capacity teacher and the smaller student. Knowledge distillation has also been used to enhance the performance of human pose estimation methods~\cite{zhao2021orderly, thai2023multiple,yang2023effective, li2021online}, including some works~\cite{thai2023multiple,zhao2021orderly} that use multiple teacher networks. 

\subsection{Cross-Dataset Human Pose Estimation}

Cross-dataset adaption has recently gained attention for 3D pose estimation~\cite{kundu2020unsupervised,gholami2022adaptpose,guan2021bilevel,zhang2020inference}. To achieve this adaption, some of these methods use generative networks~\cite{Adapting_SVCI_AAAI23}, mesh reconstruction~\cite{guan2021bilevel}, or unsupervised learning techniques~\cite{kundu2020unsupervised}. In 2D human pose estimation, cross-dataset keypoint prediction remains scarce. Some methods available in MMPose~\cite{mmpose2020}, including RTMPose~\cite{jiang2023rtmpose} and RTMO~\cite{lu2023rtmo} are trained on a "cocktail" of multiple datasets called Body7 and Body8. However, they are often only trained to predict a subset of these points instead of predicting a union of all available annotations across the datasets.

\section{Methodology}
\label{sec:methodology}

\begin{figure}
    \centering
    \includegraphics[width=\linewidth]{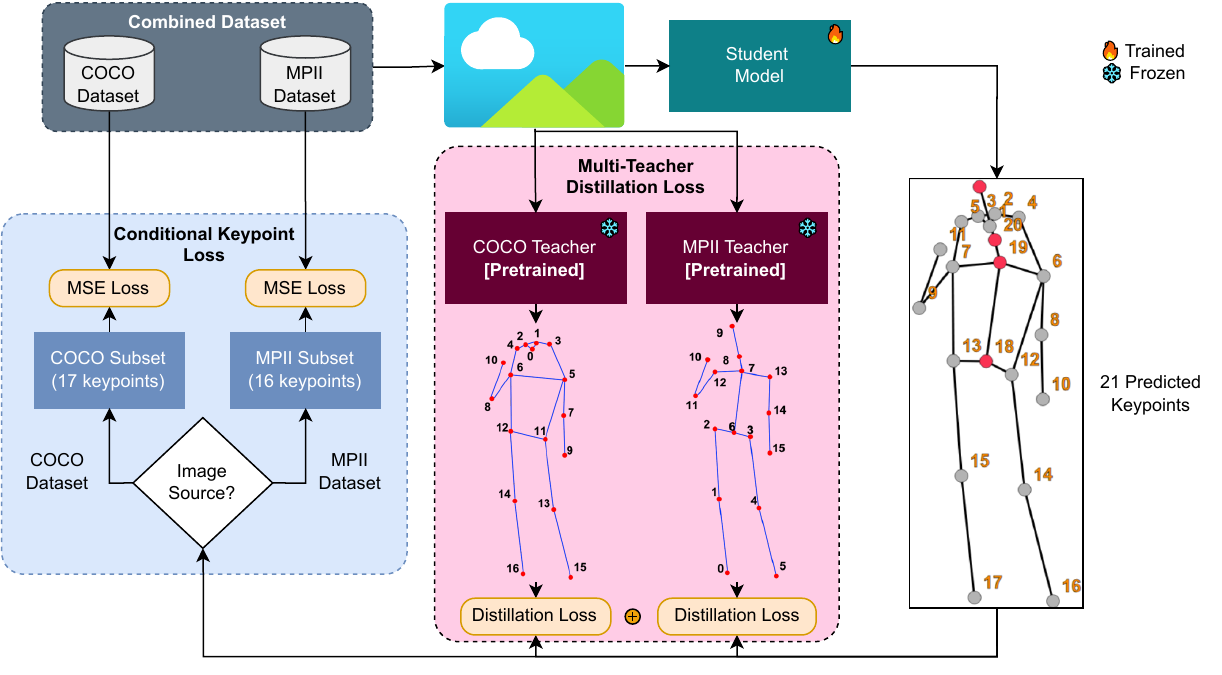}
    \caption{Illustration of our proposed learning approach, showing how information from multiple datasets is integrated to create the unified skeleton. Our student learns from both datasets and one pretrained teacher for each dataset using a combination of distillation losses and conditional keypoints loss.}
    \label{fig:methodology}
\end{figure}

\subsection{Problem Formulation}

Given a collection of keypoint datasets $\mathcal{D} = \left({D_i}\right)_{1\leq i\leq N}$, each dataset ${D}_i$ is associated with a labeled skeleton $\mathcal{K}_i$ that defines a set of keypoints. We aim to construct a superset skeleton $\mathbb{K}$, containing all unique keypoints from the individual skeletons $\mathcal{K}_i$. Formally, this is represented as:
\begin{equation}
    \mathbb{K} = \bigcup_{i=1}^{N} \mathcal{K}_i
\end{equation}
In essence, $\mathbb{K}$ includes every keypoint defined across all $N$ datasets, ensuring comprehensive pose representation.

Let $\mathcal{T} = \left({T_j}\right)_{1\leq j\leq M}$ denote the set of teacher networks, where each teacher $T_j$ is an expert on a subset of datasets and hence a subset of the superset skeleton $\mathbb{K}$. The knowledge of each teacher is encapsulated by its ability to predict keypoints for its expert datasets. Specifically, for teachers focusing on COCO and MPII datasets, the overlap and unique keypoints can be expressed as:
\begin{align}
    \mathcal{K}_{\mathrm{overlap}} &= \mathcal{K}_{\mathrm{COCO}} \cap \mathcal{K}_{\mathrm{MPII}} \\
    \mathcal{K}_{\mathrm{unique}}^{\mathrm{COCO}} &= \mathcal{K}_{\mathrm{COCO}} \backslash \mathcal{K}_{\mathrm{overlap}} \\
    \mathcal{K}_{\mathrm{unique}}^{\mathrm{MPII}} &= \mathcal{K}_{\mathrm{MPII}} \backslash \mathcal{K}_{\mathrm{overlap}}
\end{align}

The student network, denoted as $\mathcal{S}$, aims to learn the superset skeleton $\mathbb{K}$, effectively acquiring the union of knowledge from all teachers. For the COCO+MPII scenario, this learning goal translates to:
\begin{equation}
    \mathcal{K}_{\mathrm{student}} = \mathcal{K}_{\mathrm{overlap}} \cup \mathcal{K}_{\mathrm{unique}}^{\mathrm{COCO}} \cup \mathcal{K}_{\mathrm{unique}}^{\mathrm{MPII}}
\end{equation}
where $\mathcal{K}_{\mathrm{student}}$ is the set of keypoints the student learns to predict with $\left\vert\mathcal{K}_{\mathrm{student}}\right\vert = 21$.

\subsection{Loss Function}

We define our loss function as a weighted sum of the task loss, which is a conditional keypoint loss, and distillation losses for each teacher.

For an image from dataset ${D}_i$, the conditional keypoint loss, $\mathcal{L}_{CK}$, between the student's prediction $p$ and the ground truth $g$ is defined as:
\begin{equation}
    \mathcal{L}_{CK}(p,g;{D}_i) = \Sigma_{k \in \mathcal{K}_i} \left\Vert p_k - g_k \right\Vert_2^2
\end{equation}
where $p_k$ and $g_k$ are the predicted and true positions of keypoint $k$, respectively.

The distillation loss for an image, irrespective of its originating dataset, against the prediction of teacher $T_j$, is defined as:
\begin{equation}
    \mathcal{L}_D(p, T_j) = \Sigma_{k\in\mathcal{K}_{T_j}} \mathrm{KL}(p_k, t_{j,k})
\end{equation}
where $t_{j,k}$ is the prediction of keypoint $k$ by teacher $T_j$, $\mathcal{K}_{T_j}$ represents the keypoints that teacher $T_j$ can predict, and $\mathrm{KL}$ is the Kullback-Leibler divergence loss.

The total loss $\mathcal{L}$ for the student network is the weighted sum of the conditional keypoint loss and the distillation losses, respectively.
\begin{equation}
    \mathcal{L}(p,g,\{\left({T_j}\right)_{1\leq j\leq M}\};{D}_i) = \alpha \mathcal{L}_{CK}(p,g;{D}_i) + (1-\alpha)\sum_{j=1}^{M} \beta_j\mathcal{L}_D(p,T_j)
\end{equation}

Here, $\alpha$ is a weighting parameter that balances the importance of conditional keypoint loss and distillation losses, respectively, and $\beta_j$ assigns a weight for distillation loss for teacher $T_j$.

\section{Experiments}
\label{sec:experiments}

This section details the experiments we conducted to validate the effectiveness of our proposed approach, including a description of the baselines, experimental setup, and the results. We also show examples of extended annotations we created for the COCO and MPII images.
\subsection{Baselines}

In our experiments, we use various RTMPose models~\cite{jiang2023rtmpose} trained on the COCO Keypoints dataset and the MPII Human Pose dataset. We compare the networks trained with our proposed approach with their counterparts trained on individual datasets. Specifically, we report results with RTMPose-t and RTMPose-s trained on each dataset separately and then compare the same models when trained on a combined dataset using our conditional loss and multi-teacher distillation. To distill the student networks, we use pre-trained RTMPose-m models as teachers.

\subsection{Experimental Setup}

We used the pre-trained weights of RTMPose-m teachers from MMPose for both the COCO and MPII datasets. We performed various experiments where the students--RTMPose-s and RTMPose-t--were trained from scratch for 420 epochs. The training used the AdamW~\cite{loshchilov2017decoupled} optimizer with an initial learning rate of $4\mathrm{e}^{-3}$. We also used a linear learning rate scheduler with a start rate of $1\mathrm{e}^{-5}$. Additionally, we employed a cosine annealing learning rate schedule for the last 210 training epochs. The training was performed with a batch size 256, and it took 2-3 days on two V100-32GB GPUs. We followed the same data preprocessing pipeline as used by the original RTMPose models.

\subsection{Evaluation Metrics}

In our experimental analysis, we utilize two commonly accepted metrics for evaluating human pose estimation models: Percentage of Correct Keypoints (PCK) and Average Precision (AP). PCK measures the accuracy of keypoint predictions by determining the percentage of predicted keypoints that fall within a specified pixel distance from the true keypoints, effectively capturing the precision of spatial localization. AP, on the other hand, provides a comprehensive measure of precision and recall across various thresholds, assessing the model's ability to identify keypoints while penalizing false positives correctly. These metrics are instrumental in benchmarking the performance of our pose estimation models across different datasets, highlighting their precision in keypoint detection and overall detection reliability.

\begin{figure*}[ht]
\begin{center}
\includegraphics[width=\linewidth]{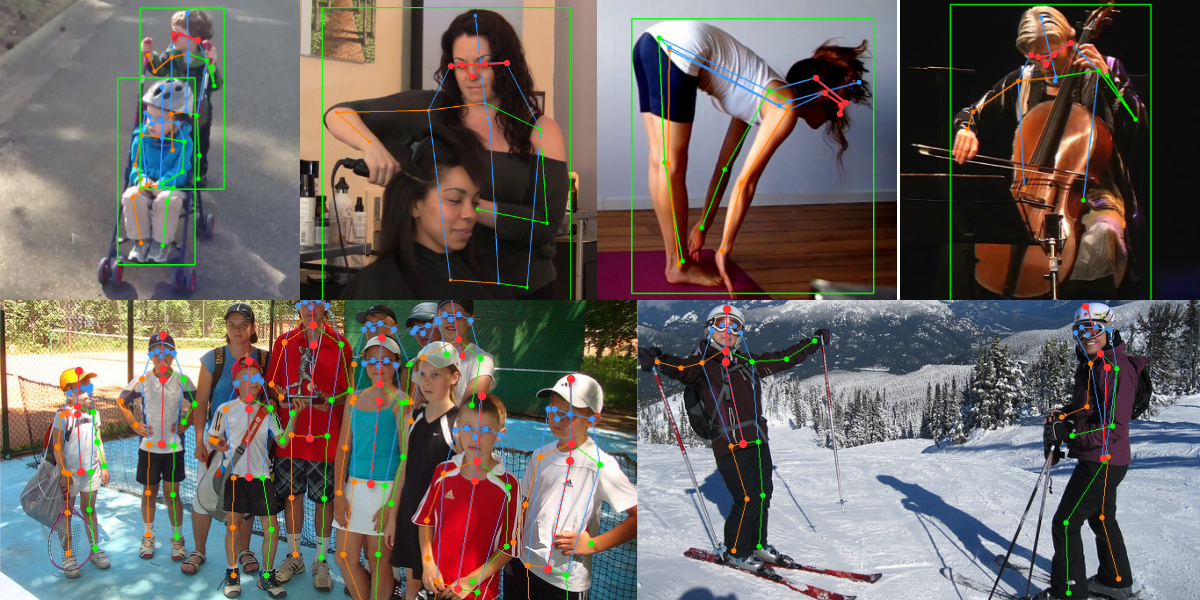}
\end{center}
   \caption{\textbf{Qualitative Results}: Predicted skeletons on different images show the accuracy of our pose estimation model across different scenarios. Notably, our model adds five new points (nose, eyes, ears) to the MPII images (top) and four new points (pelvis, thorax, neck, head top) to the COCO images (bottom). These points are highlighted in red ($\textcolor{red}{\bullet}$) in the figure.}
\label{fig:results_qualitative}
\end{figure*}

\subsection{Results}

The results from our experiments, summarized in Table~\ref{tab:results}, demonstrate the performance impacts of our unified skeleton learning approach. Although the model trained on the combined dataset using our unified learning strategy does not always surpass the performance of models trained on individual datasets in dataset-specific metrics, it demonstrates significant adaptability and broader utility.

\begin{table}[ht]
  \scriptsize
  \begin{center}
  \begin{tabular}{@{}|ll|cc|cccccc|cc|@{}}
    \hline
    & & \multicolumn{2}{c|}{MPII} & \multicolumn{6}{c|}{COCO} & & \\
    Model & Training & PCK & PCK$^{0.1}$ & AP & AP$^{0.5}$ & AP$^{0.75}$ & AR & AR$^{0.5}$ & AR$^{0.75}$ & Kpts & Avg \\
    \hline
    \hline
    \multirow{4}{*}{Tiny}  
    & MPII     & 87.75 & 29.66 & 19.82 & 57.56 &  3.93 & 25.75 & 62.37 & 14.37 & 16 & 53.79 \\
    & MPII*    & 87.75 & 29.66 & \textit{53.23} & \textit{76.61} & \textit{56.43} & \textit{59.12} & \textit{81.40} & \textit{62.46} & 16 & \textit{70.49} \\
    & COCO*    & \textit{77.62} & \textit{17.79} & 67.99 & 88.07 & 75.59 & 73.50 & 92.03 & 80.38 & 17 & \textit{\textbf{72.81}} \\
    \rowcolor{lightgray} & Unified  & 79.75 & 19.58 & 62.03 & 85.66 & 69.39 & 68.12 & 90.05 & 75.13 & \textbf{21} & \textbf{70.89} \\
    \hline
    \multirow{4}{*}{Small}
    & MPII     & 89.69 & 32.28 & 21.86 & 60.85 & 53.51 & 27.96 & 65.52 & 16.17 & 16 & 55.78 \\
    & MPII*    & 89.69 & 32.28 & \textit{57.48} & \textit{78.88} & \textit{61.57} & \textit{63.33} & \textit{83.61} & \textit{67.50} & 16 & \textit{73.59} \\
    & COCO*    & \textit{81.93} & \textit{22.82} & 72.17 & 89.19 & 79.38 & 77.23 & 92.85 & 84.74 & 17 & \textit{\textbf{77.05}} \\
    \rowcolor{lightgray} & Unified  & 84.16 & 24.66 & 68.63 & 88.30 & 75.73 & 74.14 & 92.10 & 80.71 & \textbf{21} & \textbf{76.40} \\
    \hline
  \end{tabular}
  \end{center}
  \caption{Comparative performance of the small and tiny variants of the RTMPose~\cite{jiang2023rtmpose} model on the MPII and COCO datasets under various training schemes. Models trained on a single dataset (MPII or COCO) show significant performance drops when evaluated on all keypoints of the alternate dataset, compared to evaluations on only the shared 12 keypoints (\textit{italicized values} in rows marked with *). The unified training approach (highlighted in gray) evaluates all available keypoints for the respective dataset and consistently demonstrates enhanced average performance across both datasets.}
  \label{tab:results}
\end{table}

\noindent \textbf{Dataset-Specific Performance}: Models trained solely on the MPII dataset exhibit limited generalization when applied to the COCO dataset, showing suboptimal performance metrics whether evaluated on all 17 COCO keypoints (rows 1 and 5) or only on the shared set of 12 keypoints (rows 2 and 6). Conversely, when these models are trained exclusively on the COCO dataset and assessed on MPII images for the shared keypoints (rows 3 and 7), they perform significantly better. It is important to note, however, that these evaluations exclude several critical MPII keypoints such as the pelvis, thorax, neck, and head, which are not predicted by the models trained only on COCO. In contrast, both RTMPose-s and RTMPose-t trained using our unified approach show enhanced cross-dataset generalization. This improvement underscores the robustness and adaptability of our method, effectively accommodating a more comprehensive range of keypoints across datasets.

\noindent \textbf{Extended Keypoint Prediction}: Our training strategy enables the models to predict a total of 21 keypoints, surpassing the number available in either the MPII or COCO datasets individually. This expansion is useful for applications requiring detailed pose analysis. By enabling the model to identify and process a broader array of keypoints, we significantly extend its applicability and performance in complex pose estimation scenarios.

\section{Ablation Studies}
\label{sec:ablations}

Our ablation studies evaluate the specific contributions of distillation within our framework. Moreover, we discuss the role of the conditional keypoint loss in learning a unified skeleton. 

\begin{table}[ht]
  \scriptsize
  \begin{center}
  \begin{tabular}{@{}|ll|ccc|@{}}
    \hline
    Model & Distill & MPII (PCKh) & COCO (AP) & Average \\
    \hline
    \hline
    \multirow{2}{*}{Tiny}
    & No  & \textbf{80.21} & 60.69 & 70.45 \\
    & Yes & 79.75 & \textbf{62.03} & \textbf{70.89} \\
    \hline
    \multirow{2}{*}{Small}
    & No  & \textbf{85.46} & 67.14 & 76.30 \\
    & Yes & 84.16 & \textbf{68.63} & \textbf{76.40} \\
    \hline
  \end{tabular}
  \end{center}
  \caption{Performance summary of RTMPose-t and RTMPose-s models with and without the application of distillation strategies on the MPII and COCO datasets. The table highlights the effects of distillation on overall performance metrics (PCKh for MPII and AP for COCO), showcasing the averaged results across datasets.}
  \label{tab:ablations}
\end{table}

\noindent \textbf{Role of Distillation}: The inclusion of distillation shows varied effects on model performance, which is evident in the results summarized in Table~\ref{tab:ablations}. For the "Tiny" model variant, distillation slightly reduces performance on MPII (PCKh) but improves COCO (AP) performance, resulting in an overall higher average score. Similarly, the "Small" model variant sees a minor dip in MPII performance but benefits on COCO, suggesting that distillation generally favors the COCO dataset's characteristics. These observations underscore the balancing act that distillation plays in aligning different dataset biases within a single model framework.

\begin{table}[ht]
  \scriptsize
  \begin{center}
  \begin{tabular}{@{}|ll|cccccccccc|@{}}
    \hline
    Model & Distill & AP & AP$^{0.5}$ & AP$^{0.75}$ & AP$^{M}$ & AP$^{L}$ & AR & AR$^{0.5}$ & AR$^{0.75}$ & AR$^{M}$ & AR$^{L}$ \\
    \hline
    \hline
    \multirow{2}{*}{Tiny}
    & No  & 60.69 & 85.21 & 67.71 & 58.06 & 66.13 & 67.00 & 89.77 & 73.82 & 62.67 & 73.18 \\
    & Yes & \textbf{62.03} & \textbf{85.66} & \textbf{69.39} & \textbf{59.14} & \textbf{67.73} & \textbf{68.12} & \textbf{90.05} & \textbf{75.13} & \textbf{63.69} & \textbf{74.37} \\
    \hline
    \multirow{2}{*}{Small}
    & No  & 67.14 & 87.49 & 74.67 & 64.17 & 73.13 & 72.95 & 91.78 & 79.77 & 68.62 & 79.17 \\
    & Yes & \textbf{68.63} & \textbf{88.30} & \textbf{75.73} & \textbf{65.52} & \textbf{74.79} & \textbf{74.14} & \textbf{92.10} & \textbf{80.71} & \textbf{69.91} & \textbf{80.25} \\
    \hline
  \end{tabular}
  \end{center}
  \caption{Detailed COCO dataset performance metrics for RTMPose-t and RTMPose-s models with and without distillation. This table displays a comprehensive breakdown of Average Precision (AP) and Average Recall (AR) scores across various thresholds and object sizes, illustrating the enhancement provided by distillation.}
  \label{tab:ablations_coco}
\end{table}

\noindent \textbf{COCO Dataset Specific Ablations}: In Table~\ref{tab:ablations_coco}, we look more closely at the performance metrics for the COCO dataset across various thresholds and object sizes. Distillation enhances all aspects of performance, particularly in Average Precision (AP) and Average Recall (AR) at stricter thresholds and larger object sizes. This detailed breakdown confirms that distillation not only improves general performance metrics but also enhances the model's accuracy and reliability in more challenging scenarios, such as precise localization and larger-scale object interactions.

\begin{table}[ht]
  \scriptsize
  \begin{center}
  \begin{tabular}{@{}|ll|ccccccc|cc|@{}}
    \hline
    Model & Distill & Head & Shoulder & Elbow & Wrist & Hip & Knee & Ankle & Mean & Mean@0.1 \\
    \hline
    \hline
    \multirow{2}{*}{Tiny}
    & No  & \textbf{94.51} & \textbf{91.24} & 78.61 & 69.38 & \textbf{81.34} & \textbf{72.98} & \textbf{67.69} & \textbf{80.21} & 19.51 \\
    & Yes & 93.83 & 90.86 & \textbf{78.71} & \textbf{70.14} & 79.82 & 71.77 & 66.82 & 79.75 & \textbf{19.58} \\
    \hline
    \multirow{2}{*}{Small}
    & No  & \textbf{96.01} & \textbf{94.53} & \textbf{84.51} & \textbf{77.26} & \textbf{86.33} & \textbf{80.15} & \textbf{74.87} & \textbf{85.46} & \textbf{25.33} \\
    & Yes & 95.74 & 93.38 & 83.30 & 76.80 & 83.59 & 78.02 & 73.78 & 84.16 & 24.66 \\
    \hline
  \end{tabular}
  \end{center}
  \caption{Ablation study results for RTMPose-t and RTMPose-s on the MPII dataset, detailing performance across individual body keypoints. This table compares models with and without distillation, emphasizing the effects of distillation on precision.}
  \label{tab:ablations_mpii}
\end{table}

\noindent \textbf{MPII Dataset Specific Ablations}: The impact of distillation on the MPII dataset in Table~\ref{tab:ablations_mpii} presents a nuanced picture. While the raw mean scores slightly decrease with distillation, the detailed keypoint analysis shows improvements in some keypoints (Wrist, Elbow), suggesting enhanced model sensitivity to more challenging keypoints. However, primary keypoints like Head, Shoulder, and Hip slightly decline, indicating a potential trade-off between generalizing well across datasets and maintaining top performance on dataset-specific primary keypoints.

\noindent \textbf{Role of Conditional Loss}: Our training setup uses images from two distinct datasets. This leads to situations where certain keypoints are absent in images from one dataset but present in another. The conditional loss is designed to adaptively manage this by only computing losses for available keypoints per image. This approach enables the model to learn effectively from the available data without penalizing it for missing points. It is crucial for building a robust model that can predict a comprehensive set of keypoints across datasets.

\noindent \textbf{Integration of Distillation and Conditional Loss}: The combination of distillation with our conditional loss formulation predominantly enhances overall model performance. However, it is observed that there is a slight decline in performance on the MPII dataset, where the number of images is lower compared to COCO. This could be partly attributed to the imbalance in dataset sizes and the specific setting of hyperparameters ($\beta_{MPII} = 0.25, \beta_{COCO} = 0.45, \alpha = 0.30$), which influence the degree of knowledge transfer and prioritization between datasets. Adjusting these parameters could potentially mitigate performance disparities and further optimize cross-dataset learning efficiency.

\noindent \textbf{Effectiveness of Extended Predictions} In our main experiments, we evaluated the unified model on the COCO and MPII datasets using only the subsets of keypoints specific to each dataset, excluding the newly added keypoints from evaluation. To evaluate the full capabilities of our models, including all predicted keypoints, we use the Halpe dataset~\cite{li2020pastanet}, which provides ground-truth annotations for 26 body points on COCO validation images, 20 of which align with our model's predictions. The missing thorax annotation in the Halpe dataset can be approximated using the midpoint between the shoulder points. This expanded evaluation, summarized in Table~\ref{tab:ablations_halpe}, validates our claim that our unified model can effectively predict points learned from one dataset on images in the other dataset.

\begin{table}[ht]
  \scriptsize
  \begin{center}
  \begin{tabular}{@{}|l|cccccccccc|@{}}
    \hline
    Model & AP & AP$^{0.5}$ & AP$^{0.75}$ & AP$^{M}$ & AP$^{L}$ & AR & AR$^{0.5}$ & AR$^{0.75}$ & AR$^{M}$ & AR$^{L}$ \\
    \hline
    \hline
    \multirow{1}{*}{Tiny}
    & 66.84 & 90.63 & 74.66 & 64.69 & 70.03 & 69.74 & 91.44 & 76.98 & 66.65 & 73.96 \\
    \hline
    \multirow{1}{*}{Small}
    & 72.75 & 92.69 & 80.81 & 70.35 & 76.21 & 75.20 & 93.30 & 82.17 & 72.11 & 79.48 \\
    \hline
  \end{tabular}
  \end{center}
  \caption{Performance of RTMPose models on Halpe~\cite{li2020pastanet} using GT boxes, demonstrating the effectiveness of our unified approach in accurately predicting an extended keypoint set.}
  \label{tab:ablations_halpe}
\end{table}

\section{Conclusion}
\label{sec:conclusions}

This paper introduced a comprehensive framework for enhancing human pose estimation by unifying keypoint annotations from multiple datasets using a multi-teacher distillation approach. Our model's ability to generalize across datasets without compromising accuracy represents a significant advancement in pose estimation technologies. Specifically, the model's capacity to predict an extended set of keypoints demonstrates its potential in applications requiring detailed human pose analysis.

Key contributions of this work include the development of a unified model that outperforms traditional single-dataset models in cross-dataset settings and the detailed experimental analysis that validates the effectiveness of our approach. Future work will explore refining these techniques, potentially using this method to extend ground-truth annotations for training sets of the two datasets, and examining active learning strategies to make the annotation process more precise.

\bibliography{egbib}
\end{document}